\pdfoutput=1

\documentclass[11pt]{article}

\usepackage[]{emnlp2024-demo}

\usepackage{float}
\usepackage{verbatim}
\usepackage{times}
\usepackage{latexsym}
\usepackage{helvet} 
\usepackage{courier}  
\usepackage{url}  
\usepackage{graphicx} 
\usepackage{natbib}  
\usepackage{caption} 
\usepackage{hyperref}
\usepackage[T1]{fontenc}

\usepackage[utf8]{inputenc}

\usepackage{microtype}
\usepackage[font=small,skip=0pt]{caption}
\usepackage{fontawesome5}

\newcommand*{\affmark}[1][*]{\textsuperscript{#1}}

\usepackage{algcompatible}
\usepackage{algorithm}
\usepackage{booktabs}
\usepackage{array}
\usepackage{makecell}

\title{DriveThru: a Document Extraction Platform and Benchmark Datasets for Indonesian Local Language Archives}

\author{
Mohammad Rifqi Farhansyah\affmark[1]\thanks{\ \ \ Contributed equally}, \ \ 
Muhammad Zuhdi Fikri Johari\affmark[2]\footnotemark[1], \ \ 
Afinzaki Amiral \affmark[3], \ \ \\
\textbf{Ayu Purwarianti\affmark[1],} \ \ 
\textbf{Kumara Ari Yuana\affmark[2],} \ \ 
\textbf{Derry Tanti Wijaya}\affmark[4]\thanks{\ \ \ Corresponding author} \\
 Institut Teknologi Bandung\affmark[1] \ \ 
 Universitas Amikom Yogyakarta\affmark[2] \\ 
 Universitas Dian Nuswantoro\affmark[3] \ \ 
 Boston University\affmark[4] \\
 \texttt{wijaya@bu.edu} \\
 }
 
\begin{document}
\maketitle
\begin{abstract}
Indonesia is one of the  most diverse countries linguistically. However, despite this linguistic diversity, Indonesian languages remain underrepresented in Natural Language Processing (NLP) research and technologies. In the past two years, several efforts have been conducted to construct NLP resources for Indonesian languages. However, most of these efforts have been focused on creating manual resources thus difficult to scale to more languages. Although many Indonesian languages do not have a web presence, locally there are resources that document these languages well in printed forms such as books, magazines, and newspapers. Digitizing these existing resources will enable scaling of Indonesian language resource construction to many more languages. In this paper, we propose an alternative method of creating datasets by digitizing documents, which have not previously been used to build digital language resources in Indonesia. DriveThru is a platform for extracting document content utilizing Optical Character Recognition (OCR) techniques in its system to provide language resource building with less manual effort and cost. This paper also studies the utility of current state-of-the-art LLM for post-OCR correction to show the capability of increasing the character accuracy rate (CAR) and word accuracy rate (WAR) compared to off-the-shelf OCR. The platform is available online at \url{https://ocrdt.ragambahasa.id/} and the benchmark dataset, evaluation script, and the models are available at our GitHub repository\footnote{\url{https://github.com/ragambahasa}}. We also provide a short ($\sim$1-minute) screencast of our system on YouTube: \url{https://youtu.be/q5uJOHKcBsg}

\end{abstract}
\section{Introduction}
Indonesia is known as one of the world’s most populated countries with a population exceeding 270 million. Spreading over 17 thousand islands, Indonesia is also one of the most diverse countries in the world with over 1,300 ethnic groups speaking over 700 local languages \cite{ethonologue27ed2024}. In terms of the number of speakers, the top 20 of Indonesian languages are spoken by over 1 million people \textit{each}.  Despite this linguistic diversity, Indonesian languages remain underrepresented in Natural Language Processing (NLP) research and technologies \cite{aji-etal-2022-one}. Thus, very little of NLP’s significant progress in the past few years have found its way to applications for these languages. 

To spur the development of research and technologies for Indonesian languages, in the past two years several efforts have been carried out to construct NLP resources for Indonesian languages \cite{cahyawijaya-etal-2023-nusacrowd,winata-etal-2023-nusax,cahyawijaya-etal-2023-nusawrites}. However, these efforts have been focused on a small number of Indonesian languages (top 10 languages in terms of existing Web presence). The resources are created either manually or translated from English resources using existing machine translation (MT) systems. As the process of hiring annotators and managing annotation is costly and time-consuming, and because existing MT systems are limited in coverage, these efforts are difficult to scale to more languages. 


Although many Indonesian languages do not have a web presence, locally there are resources that document these languages well in printed forms such as books: textbooks, grammar books, dictionaries, story books, etc., magazines, and newspapers. Digitizing these existing resources will enable scaling of Indonesian language resource construction to many more languages. In addition, digitizing existing resources such as books can alleviate some of the drawbacks of prior works. 
Firstly, a book must have passed through quality assurance stages before being published, thus, the requirement of recruiting native speakers as a dataset curator can be altered since the works will be mostly focused on collecting books and identifying languages of these books instead of creating resources from scratch. 
Secondly, the time and cost for constructing resources can be reduced since it takes less time and costs to digitize books than creating resources from scratch. Several books are available online and published openly by the Indonesian National Library or the Indonesian government itself. 

In this study, we propose \textbf{DriveThru} platform, an alternative system that digitizes documents to assist Indonesian NLP researchers in their language resource collection. 
In this work, aside from using an off-the-shelf OCR system, TesseractOCR, we also benchmark LLMs for post-OCR error correction in local Indonesian languages, which occurs when off-the-shelf OCR systems are unable to recognize certain characters. 
We explore OCR and post-OCR error correction four low-resource Indonesian local languages: Javanese (jav), Sundanese (sun), Minangkabau (min), and Balinese (ban) written in latin scripts. This work demonstrates our approach for collecting underrepresented language resources through document extraction i.e., digitization of printed documents written in these languages. 

\section{Related Work}
The development of a language resource dataset via document extraction encounters significant challenges including the presence of noisy data, incorrect character recognition, and frequent occurrences of hallucinations. Several studies on Finnish language encountered challenges when they had to digitize old documents with previously unseen fonts \cite{Drobac_Kauppinen_Lindén_2017, Koistinen_Kettunen_Pääkkönen_2017}.

Prior works have attempted to address these issues by implementing post-processing of the OCR outputs. Many prior works involve manual corrections of the OCR outputs. For example, to develop language resources for the Bodo language, one of the community languages spoken in India \cite{narzary-etal-2022-generating}, the project utilizes Google Docs for annotators to manually correct OCR outputs. Another work \cite{clematide-etal-2016-crowdsourcing} developed a crowd-correction platform called Kokos to improve the quality of OCR outputs by engaging volunteers to correct digitized yearbooks written in German and French. Although manual corrections are simple, it can be time-consuming and costly as it requires recruiting and training many native speakers of the language to be involved to produce numerous datasets. 

Other prior works have attempted to overcome some of the shortcomings of these prior works by implementing an automation process for OCR post-correction. When there is no or limited dataset available for a language, using an OCR system's output to post-correct another OCR system's output by comparing the two can be an alternative. This has been done for post-OCR text correction in romanized Sanskrit \cite{Krishna_Majumder_Bhat_Goyal_2018}. Other work \cite{Poncelas_Aboomar_Buts_Hadley_Way_2020} has created a tool for correcting common errors in the Tesseract OCR engine for an English chapter book. These approaches however, can only be conducted by people who know how to program because the tool interface is in command-line format.

In addition, large language models (LLMs) have recently been employed to conduct automatic OCR post-correction. Pre-trained models such as ByT5 \cite{Löfgren_Dannélls_2024}, which operates at the character level, have been utilized for Swedish. Furthermore, Google Vision AI toolkit combined with LSTM has been employed for endangered languages such as Ainu, Griko, and Yakka \cite{Rijhwani_Anastasopoulos_Neubig_2020}. Additionally, a fully unsupervised character-based sequence-to-sequence NMT model has been applied for error correction in English and Finnish \cite{Duong_Hämäläinen_Hengchen_2021}. State-of-the-art systems for OCR post-correction involves generative LLMs and the use of prompt-based approaches using LLAMA2 to successfully reduce character error rate for English newspapers \cite{Thomas_Gaizauskas_Lu_2024}. 

Several of these studies have also integrated OCR with web applications. However, many of these applications are trained in high-resource languages \cite{cassidy-2016-publishing} such as English. Additionally, some projects are no longer maintained \cite{reynaert-2014-ticclops}, rendering them inaccessible \cite{weerasinghe-etal-2008-nlp}. In this work, we develop a platform called DriveThru for low-resource language document extraction and study the effectiveness of automatic OCR post-corrections for these languages.

\section{System Description}
\begin{figure}[h]
    \centering
    \includegraphics[width=0.75\linewidth]{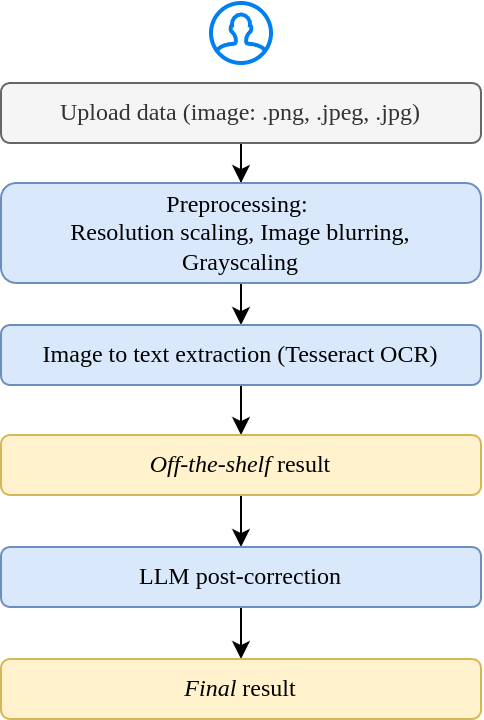}
   \vspace{10pt}
    \caption{The figure shows how the platform's extraction mechanisms work behind the scenes after users upload their images. The first step involves preprocessing the images. Then, Tesseract OCR extracts the text, producing the initial output. Finally, a language model will perform text correction to get the final results of document extraction.}
    \label{fig:system-workflow}
\end{figure}

DriveThru platform is inspired by the concept of fast-food restaurant drive-thru services, which allow customers to order food without leaving their vehicle, entering the restaurant, and ordering from a waitress. We use this philosophy in our system so that users do not need to create an account or log in to the apps; instead, they simply upload their document in the form of images (e.g., scans of printed documents). 
Finally, they can get the extracted text after the entire process takes place inside the DriveThru platform. The system workflow is shown in Figure \ref{fig:system-workflow}. 

DriveThru accepts several image formats including: \texttt{.png, .jpg, 
} or \texttt{.jpeg}. The maximum image uploaded in one cycle is 5 images. Any input surpassing this limit will be rejected by our system. The screen capture of DriveThru's user interface is provided in Figure \ref{fig:drive-thru-UI} in the Appendix. 

\subsection{Vocabulary Dataset}
\label{sec:vocabulary-dataset}
To construct one of our post-OCR correction model (i.e., the LLM's \textit{few-shot prompting} approach where we prompt LLM to correct an OCR output, providing potentially relevant words from dictionaries), we utilize a \textbf{word dictionary} that consists of paired vocabulary entries in bahasa Indonesia (i.e., Indonesian) and the Indonesian low-resource local language. While most of these entries are single words, some include multi-word expressions. These word pairs are extracted from dictionary books of the low-resource language (Table \ref{tab:dataset_numbers}). 
\begin{table}
\centering
\begin{tabular}{|c|c|}
    \hline
    \textbf{Language} & \textbf{Number of Pairs} \\
    \hline
    Sunda & 10831 \\
    \hline
    Jawa & 14680 \\
    \hline
    Minang & 12503 \\
    \hline
    Bali & 45120 \\
    \hline
    \end{tabular}
    \caption{Number of vocabulary pairs for each language used in the training dataset.}
    \label{tab:dataset_numbers}
\end{table}

\subsection{Similar Words}
\label{sec:similar-words}
The previously collected vocabulary dataset is instrumental in identifying the most similar words for each token in the OCR output to be corrected. We use the Longest Common Substring (LCS) algorithm that determines the longest sequence of shared characters between two words to measure their similarity effectively.


Following the similarity computation, we undertake a selection phase to refine the list of similar words for each token in the input text (i.e., the OCR output to be corrected). This selection process is structured as follows:
    \begin{enumerate}
    \item \textbf{Similarity Assessment:} Each token in the input text is compared to every word in the word dictionary using the LCS algorithm. Words that achieve a similarity score above a specified threshold are identified for further consideration.
    \item \textbf{Relevance Filtering:} Words that display excessive similarity (more than $K$ entries) are removed from the list, ensuring that only the most relevant matches are retained for analysis.
    \item \textbf{Optimized Selection:} To maintain efficiency and manage prompt length, the number of similar word pairs is capped at a maximum of 10. When the number of relevant pairs exceeds this limit, a random sampling method is used to finalize the selection.
\end{enumerate}
This systematic approach not only improves the precision of our post-OCR corrections but also guarantees that the process remains both efficient and scalable.


\subsection{Benchmark Dataset}
To evaluate our OCR post-correction, we leverage books, manuscripts, and magazines 
obtained from the National Library Of Indonesia\footnote{\url{https://www.perpusnas.go.id}} and the Indonesian Ministry of Education, Culture, Research, and Technology (MoECRT) repository websites\footnote{\url{https://repositori.kemdikbud.go.id}}. 
Documents obtained from those websites are following license states in Appendix \ref{appx:copyright-license}. The document titles that we use to evaluate our OCR post-correction approach are listed in the Table \ref{tab:dataset} in the Appendix. 

\subsection{Preprocessing}
The DriveThru workflow starts with preprocessing of the uploaded image file, see Algorithm \ref{alg:image-scaling}. We use the OpenCV\footnote{\url{https://github.com/opencv/opencv}} (cv2) python library in this work and begin by scaling the image resolutions times 1024 pixels if the uploaded image width is less than 1024 pixels. This is to make the processed image resolutions larger than 1024 pixels square and ensure that the image's content does not disappear during the next preprocessing step. We use \textit{\texttt{cv2.INTER\_CUBIC}} modules since it's the only modules that produce high-quality and sharply focused images in our case. 

\begin{algorithm}[]
\caption{Image resolution scaling}\label{alg:image-scaling}
\begin{algorithmic}
\REQUIRE $w \gets 1024$
\STATE $W \gets w$

\IF {$W < 1024$} 
    \STATE $ratio = 1024 / W$
    \STATE $image = cv2.resize(image$, 
                        \STATE \hspace*{2em}$fx=ratio, fy=ratio, $
                        \STATE \hspace*{2em}$interpolation=$  \textbackslash
                        \STATE \hspace*{2em}$cv2.INTER\_CUBIC$)
\ENDIF 

\STATE $gray = cv2.cvtColor(image,$
                                \STATE \hspace*{3em} $cv2.COLOR\_BGR2GRAY)$
\STATE $blur = cv2.GaussianBlur(gray, (3,3), 0)$

\STATE $thresh = cv2.threshold(blur, 0, 255, $
            \STATE \hspace*{2em} $cv2.THRESH\_BINARY\_INV + \textbackslash$
            \STATE \hspace*{2em} $cv2.THRESH\_OTSU)[1]$

\STATE $invert = 255-thresh$

\end{algorithmic}
\end{algorithm}

In addition, we apply the conversion of image to grayscale using \textit{\texttt{cv2.COLOR\_BGR2GRAY}} module to prevent the OCR engines from being affected by color distractions. The next preprocessing step involves a blurring effect on the image using \textit{\texttt{cv2.GaussianBlur}}, which is essential for eliminating any background noise. To remove noise from the image, we apply a threshold using \textit{\texttt{cv2.threshold}} to retain the densely packed pixels and discard the sparse ones. Finally, we invert the image colors from black-to-white to white-to-black by subtracting the threshold value from the 255 RGB value.

\subsection{Model}
\subsubsection*{OCR Engines (TesseractOCR)}
The image-to-text process is conducted using TesseractOCR\footnote{\url{https://github.com/tesseract-ocr}} without any fine-tuning of the base model. To achieve the most accurate extraction results compared to the ground truth, adjustments are made by configuring the OCR Engine mode (\textit{\texttt{--oem}}) to 3 (default settings based on engine availability) and the Page Segmentation Mode \textit{(\texttt{--psm}}) to 6. 
The language setting is left at its default, English, as the low-resource language documents we are extracting are written in Latin scripts. 

To integrate TesseractOCR with a Python-based web application, we utilized PyTesseract\footnote{\url{https://github.com/h/pytesseract}} as an interface for the TesseractOCR within the Python programming environment. The output from the off-the-shelf OCR is subsequently processed through LLM post-correction using 3 distinct LLMs, which will be detailed in the following sections.

\subsubsection*{LLaMA 3}
Llama 3\footnote{\url{https://github.com/meta-llama/llama3}} or Meta Llama 3 is a successor of Llama 2 \cite{touvron2023llama2openfoundation} which is a group of open-source pre-trained and instruction-tuned generative text models made by Meta AI \cite{llama3modelcard}. It was released in two parameter sizes, 8B and 70B, both in pre-trained and instruction-tuned types. Llama 3 was pre-trained with over 15 trillion tokens from public data. The fine-tuning used public instruction datasets and more than 10 million human-annotated examples. We choose Llama 3 as it has the best accuracy as well as performance among existing LLM model that is pre-trained on English-centric data. This project uses \href{https://huggingface.co/meta-llama/Meta-Llama-3-70B-Instruct}{Meta-Llama-3-70B-Instruct} model for 
our post-OCR correction model. This model is employed in both zero-shot and few-shot prompting scenarios to correct misspelled words in Indonesian local languages. By integrating this advanced model, we effectively address spelling errors in local languages, demonstrating the versatility and robustness of our approach.

\subsubsection*{GPT-4}
GPT-4, the successor to GPT-3.5, is a large-scale and multimodal model capable of handling both text and image inputs and generating text outputs \cite{openai2024gpt4technicalreport}. It surpasses human performance in various professional and academic assessments. This large language model excels in bug improvement and reinforces foundational knowledge from previous models and most state-of-the-art systems. GPT-4 demonstrates superior English-language performance compared to other large language models, including its predecessor GPT-3.5, and also performs well in low-resource languages such as Latvian, Welsh, and Swahili. However, GPT-4 retains some limitations similar to previous models, including the occurrence of hallucinations, a limited context window, and an inability to learn from previous events. We employ GPT-4 for post-correction of OCR outputs due to its pre-training with the Indonesian language. Similar to the previous model, this model is utilized in both zero-shot and few-shot prompting contexts to correct misspelled words in Indonesian indigenous languages.

\subsubsection*{Post-Correction Approach}
\begin{figure}[h]
    \centering
    \includegraphics[width=1\linewidth]{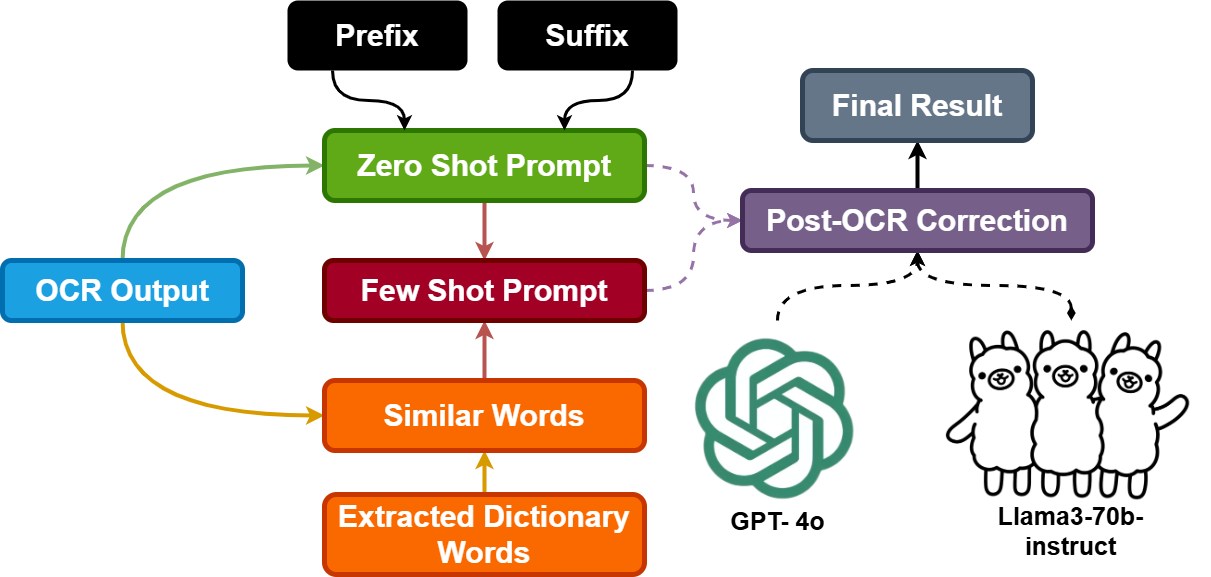}
   \vspace{10pt}
    \caption{The figure shows our post-OCR correction flow with 2 approach: few-shot and zero-shot prompting.}
    \label{fig:postcorrectionocr-workflow}
\end{figure}
As outlined above, this project employs two advanced techniques for post-OCR correction: zero-shot and few-shot prompting. These techniques are implemented as follows (Figure \ref{fig:postcorrectionocr-workflow}):

\begin{enumerate}
    \item \textbf{Vocabulary Dataset Collection:} As described in Section \ref{sec:vocabulary-dataset}, we collect vocabulary datasets for each local language. These datasets are essential for identifying the most relevant similar words for each token in the input text i.e., the OCR output to be corrected.

    \item \textbf{Similar Words Formation:} We form a list of similar words by comparing the words in the dictionary with each token in the input text, focusing on those with the highest similarity scores. For a detailed explanation of the similarity calculation mechanism, refer to Section \ref{sec:similar-words}.

    \item \textbf{Prompt Construction:} After obtaining the relevant data, we construct a prompt that includes a prefix (i.e., "\textit{Fix the grammar of the following text}"), suffix (i.e., "\textit{The following are potentially similar words from the dictionary}"), and the input text (i.e., OCR output) itself. In zero-shot prompting, the prefix and input text are directly input into the Large Language Model (LLM) as an instruction to execute. In contrast, for few-shot prompting, the prompt is enhanced with the suffix that contains additional hints in the form of a list of word pairs from the input text and similar vocabulary from the dictionary.

    \item \textbf{Result Generation:} The final output is generated based on the responses provided by the LLM. This output incorporates the corrections suggested by the model based on the input and the  prompt.
\end{enumerate}
\section{Results and Discussion}
\begin{table*}[ht]
\small
\centering
\begin{tabular}{@{}rlrrrrr@{}}
\toprule
\multicolumn{1}{l}{} &                   & \multicolumn{5}{c}{\textbf{CAR}}                                                                                                                                                                           \\ \midrule
\multicolumn{1}{l}{} & \textbf{Language} & \multicolumn{1}{c}{\textbf{OTS}} & \multicolumn{1}{c}{\textbf{Llama3 (ZS)}} & \multicolumn{1}{c}{\textbf{Llama3 (FS)}} & \multicolumn{1}{c}{\textbf{GPT-4 (ZS)}} & \multicolumn{1}{c}{\textbf{GPT-4 (FS)}} \\
\midrule
1                    & Balinese          & 0.943                            & 0.917                                    & 0.919                                    & 0.893                                   & 0.914                                   \\
2                    & Javanese          & -0.993                           & 0.970                                    & 0.956                                    & 0.965                                   & 0.965                                   \\
3                    & Sundanese         & 0.911                            & 0.738                                    & 0.168                                    & -0.368                                  & -0.699                                  \\
4                    & Minangkabau       & 0.958                            & 0.942                                    & 0.942                                    & 0.924                                   & 0.926                                   \\
\midrule
\multicolumn{2}{c}{\textbf{avg (\%)}}    & 0.45475                          & \textbf{0.892}& 0.746                                    & 0.603                                   & 0.526                                   \\ \bottomrule
\end{tabular}
\caption{Percentage of Character Accuracy Rates (CAR) on different extraction techniques. The off-the-shelf (OTS) column means there are no additional steps to repair the extracted text, while the others perform additional post-correction steps involving LLMs. Overall, Llama3 with a Zero-shot approach outperforms other LLMs in post-correction OCR.}
\label{tab:car-eval}
\end{table*}

DriveThru was developed to assist future NLP researchers, students, scholars, data scientists, organizations, language enthusiasts, and other entities with their language resource collection tasks. We aim to make the system's interface as simple as possible, even if no instructions are provided in beforehand. 

Using DriveThru is straightforward. As shown in Figure \ref{fig:drive-thru-UI} in the Appendix, users can upload files by simply dragging and dropping them from their file managers into the designated area, or if users prefer to browse their file through the "browse files" dialog boxes it can be achieved by clicking on the drag-and-drop area. The platform allows for the upload of up to five files simultaneously. If a file is mistakenly uploaded, it can be removed by clicking the cross icon below the drag-and-drop area, or by selecting the light gray "Clear" button to remove all entries. To proceed with the uploaded files, users can click the red "Proceed" button.

\subsection*{Evaluation}
The off-the-shelf (OTS) TesseractOCR engine is considered capable of recognizing the majority of the provided images, though some hallucinations occur. As shown in Table \ref{tab:maxlen-eval} in the Appendix, in terms of word counts, OTS Tesseract produces a higher total word count than the ground truth (GT), with a difference of 2,699 words. According to the CAR in Table \ref{tab:car-eval} and WAR scores in Table \ref{tab:war-eval} in the Appendix, OTS Tesseract demonstrates the highest accuracy. However, it struggles with detecting images containing Javanese document archives, resulting in the lowest score among the other languages.

Applying zero-shot learning for post-correction OCR improves the average word and character accuracy of Llama3 models. Even yet, it appears that the Balinese, Sundanese, and Minangkabau scores do not differ significantly from the OTS scores. In contrast, for Javanese, there was a significant improvement from below zero percent to above 50 percent, leading to fewer hallucinations than before.

Even if the scores for some approaches (either zero-shot or few-shot) are better than the OTS, it may not be sufficient. This is because in our human annotation file, all entries are written as they appear, regardless of whether they are correct or not by \textit{lexical rules}. Post-OCR correction using zero-shot or few-shot techniques can achieve more by fixing punctuation, hyphenated words, removing unclear parts that were misinterpreted by OCR, and more. However, if the OCR output is severely distorted, post-OCR correction still faces significant challenges.

From the text above, further details can be illustrated through qualitative examples in Table \ref{tab:ocr_comparison} in the Appendix:

1. \textbf{Example 1} in Table \ref{tab:ocr_comparison} shows that post-OCR correction can resolve the hyphenated word problem when scanning a document using OCR. For instance, "ndu- weni" is replaced by "nduweni", "ba- nget" is replaced by "banget", and "ma- mah" is replaced by "mamah".

2. \textbf{Example 2} in Table \ref{tab:ocr_comparison} demonstrates that post-OCR correction can remove unclear parts from the image that have been scanned with OCR if they are not detected as part of the language vocabulary. It removes "Bataan ceeiaalteleea Saati: St meinen, aa" because the post-OCR correction detects that this string is merely an unclear scanned part from the image and is not included in the language vocabulary.

3. \textbf{Example 3} in Table \ref{tab:ocr_comparison} illustrates that post-OCR correction can also solve problems related to punctuation usage. The post-OCR correction can remove unclear parts from the text and add a period at the end of the sentence.

\subsection*{Limitations}
The RagamBahasa platform has not yet been integrated with a post-processing model due to a lack of independent computing resources for the project. Currently, the models are temporarily deployed on high-performance computing resources of an academic institution, which are shared with other research projects. We have not applied enhancements to the OCR engine to reduce character misinterpretation during text recognition. While it is adequate for extracting Indonesian local languages in Latin script, improvements are needed for reading regional scripts (e.g., Javanese, Balinese, Sundanese, etc.). Our benchmarks have only been demonstrated on languages classified as Institutional-Stable by Ethnologue. Despite being low-resource languages, scanned language resources are relatively accessible online. This effort should be expanded to include languages classified as endangered, such as Betawi, Acehnese, and Wolio, with more diverse capturing techniques. Regardless of the limitations of our tools, this study meets our requirements for building a language resource database of Indonesian local languages in the future.

In addition, as illustrated in Example 4 from Table \ref{tab:ocr_comparison}, when the OCR output is highly distorted or unclear, it becomes exceedingly difficult to accurately identify the actual text, even with advanced LLMs using zero-shot or few-shot prompting. This limitation underscores the challenge of dealing with severely degraded OCR input and highlights the need for further improvements in both OCR technology and post-processing techniques to handle such cases more effectively.

\section*{Ethics Statement}

The primary objective of this project is to utilize scanned language resources of Indonesian local languages available online in the Indonesian government archives repository. This initiative is also funded by the same governmental entity to ensure alignment between the research objectives and the funding body. All datasets and source code used in this study are publicly accessible, with copyright and licensing details duly specified.

\section*{Acknowledgments}
We are grateful for the financial support and technical assistance on this research provided by the Indonesian Ministry of Education, Culture, Research, and Technology (MoECRT) and the Indonesia Endowment Fund for Education (LPDP) through the ACE Open Research program, part of the US-Indonesia collaboration program. We would like to thank Boston University for providing computing facilities through Shared Computing Cluster (SCC) and Monash University Indonesia for the essential to the success of this project.

\bibliography{anthology,main}

\appendix
\section{Copyright and License \label{appx:copyright-license}}

In addition to following the author's license, we follow the Indonesian MoECRT repository policy\footnote{\url{https://repositori.kemdikbud.go.id/information.html}} which states "\textit{..pengguna yang menggunakan sumber informasi dari Repositori Institusi Kemendikbudristek harus menuliskan atribut sumber yang digunakan dan tidak digunakan untuk tujuan komersial.}" This means that anyone who uses resources from the repositories must acknowledge the source and not commercialize the work when disseminating it. 

Align with the consideration, the content produced by DriveThru is licensed under CC-BY NC 4.0, which means that people can use our platform to create dataset they need without commercializing or disrespecting the authors' work.


\begin{figure}[h]
    \centering
    \includegraphics[width=1\linewidth]{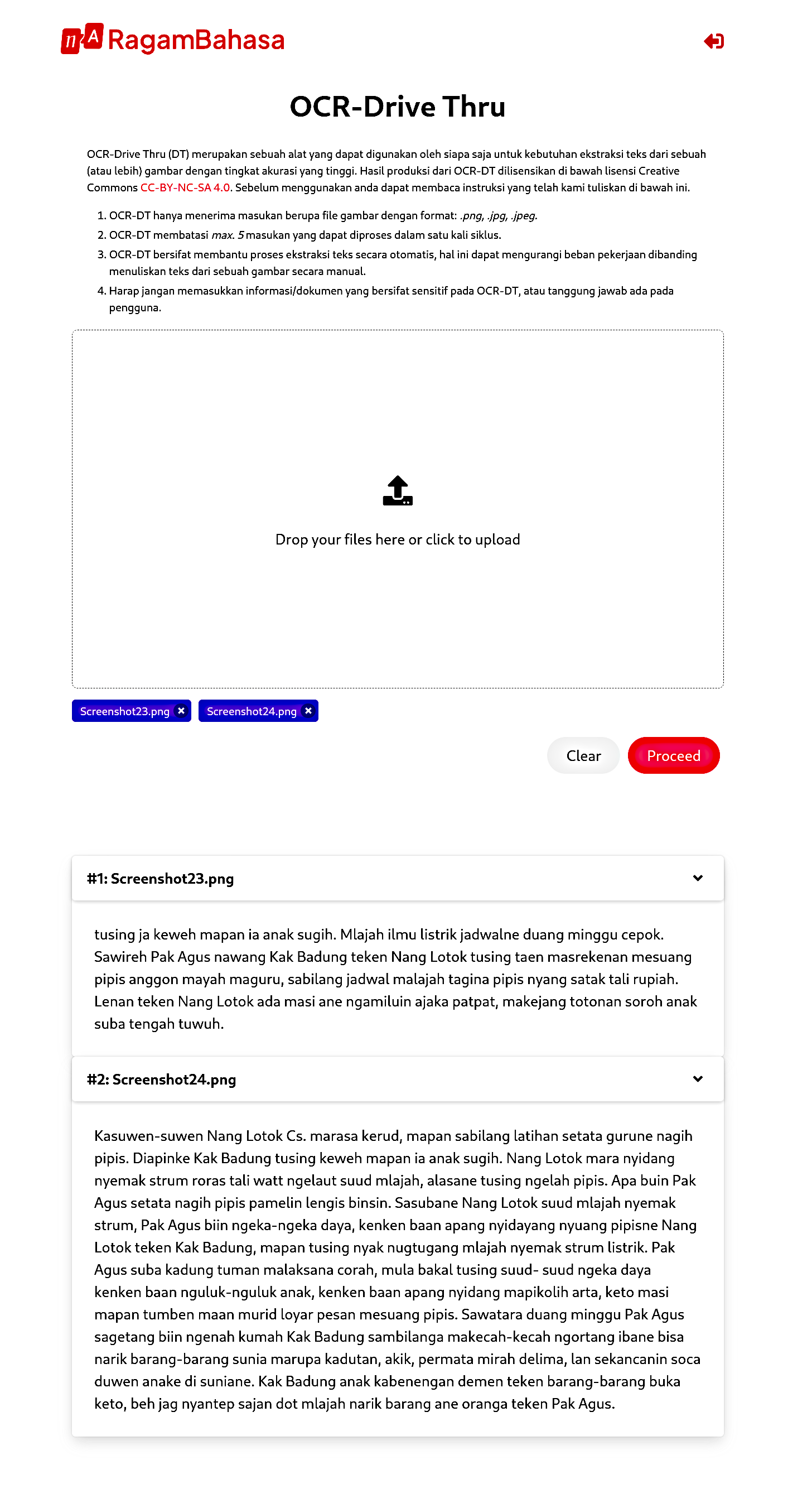}
    \caption{Screen capture of DriveThru application interface, It shows the instruction to be considered below the application title, it also shows how the application previews two OCR outputs of the uploaded files below the drag-and-drop area}
    \label{fig:drive-thru-UI}
\end{figure}

\begin{table*}[]
\centering
\begin{tabular}{@{}rlllll@{}}
\toprule
\multicolumn{1}{l}{\textbf{\#}} & \textbf{Language} & \textbf{ISO} & \textbf{Source Title}                             & \textbf{Genre} & \multicolumn{1}{l}{\textbf{\makecell[l]{Total of\\Image}}} \\ \midrule
1                               & Balinese          & ban          & I Bagus Caratan                                   & Children       & 26                                          \\
2                               & Balinese          & ban          & Leak Pemoroan                                     & Short Story    & 24                                          \\
3                               & Balinese          & ban          & \makecell[l]{Ejaan Bahasa Daerah Bali\\yang Disempurnakan 1974}& Learning Book  & 13                                          \\
4                               & Balinese          & ban          & Majalah Suara Saking Bali Edisi VII               & Magazine       & 12                                          \\
5                               & Balinese          & ban          & Paparikan Lawe                                    & Learning Book  & 10                                          \\
6                               & Balinese          & ban          & \makecell[l]{Pedoman Umum Ejaan Bahasa Bali\\Dengan Huruf Latin}& Learning Book  & 15                                          \\
7                               & Javanese          & jav          & Panjebar Semangat                                 & Magazine       & 100                                         \\
8                               & Sundanese         & sun          & Carita ti Carita                                  & Short story    & 5                                           \\
9                               & Sundanese         & sun          & Hayam Gecok Ngeunah                               & Short story    & 5                                           \\
10                              & Sundanese         & sun          & Mangle                                            & Magazine       & 25                                          \\
 11& Sundanese& sun& Wawacan Rengganis & Story Book  &16\\
 12& Sundanese& sun& Wawacan Sejarah Anbia & Story book  &10\\
 13& Sundanese& sun& Raja Neger Jeung Bangsa Arab & Story book  &17\\
 14& Sundanese& sun& Lain Eta & Story book  &12\\
 15& Sundanese& sun& Istri kasasar & Story book  &10\\
16& Minangkabau       & min          & \makecell[l]{Kaba Bujang Paman\\Dan Kaba Rambun Pamenan}         & Novel          & 55                                          \\
17& Minangkabau       & min          & \makecell[l]{Kaba Kambang Luari Sutan\\Pangaduan}                & Novel          & 45                                          \\
\midrule
\multicolumn{5}{c}{\textbf{Total}}                                                                                                      & \textbf{400}\\ \bottomrule
\end{tabular}
\caption{A list of document titles from four different languages with their respective genres. We collected 335 images from a total of 12 books spanning various genres in language resource archives to use for benchmark datasets.}
\label{tab:dataset}
\vspace{100mm}
\end{table*}

\begin{table*}[t]
    \centering
    \begin{tabular}{|m{0.18\textwidth}|m{0.22\textwidth}|m{0.22\textwidth}|m{0.22\textwidth}|}
        \hline
        \textbf{Image} & \textbf{OCR Output} & \textbf{Human Annotation} & \textbf{Post-OCR Correction} \\
        \hline
        \includegraphics[width=0.2\textwidth]{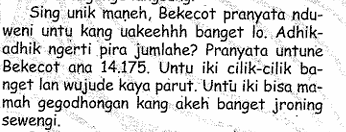} & 
        Sing unik maneh, Bekecot pranyata ndu- weni untu kang uakeehhh banget lo. Adhik- adhik ngerti pira jumlahe? Pranyata untune Bekecot ana 14.175. Untu iki cilik-cilik ba- nget lan wujude kaya parut. Untu iki bisa ma- mah gegodhongan kang akeh banget jroning sewengi. : : & 
        Sing unik maneh, Bekecot pranyata ndu- weni untu kang akeh banget lo. Adhik- adhik ngerti pira jumlahé? Pranyata untune Bekecot ana 14.175. Untu iki cilik-cilik ba- nget lan wujude kaya parut. Untu iki bisa ma- mah gegodhongan kang akeh banget jroning sewengi. & 
        Sing unik maneh, Bekicot pranyata nduweni untu kang uakeh banget lo. Adhik-adhik ngerti pira jumlahe? Pranyata untune Bekicot ana 14.175. Untu iki cilik-cilik banget lan wujude kaya parut. Untu iki bisa mah gegodhongan kang akeh banget jroning sewengi. \\
        \hline
        \includegraphics[width=0.18\textwidth]{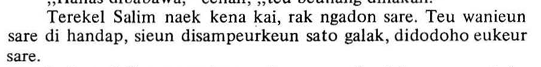} & 
        Bataan ceeiaalteleea Saati: St meinen, ~ aa Terekel Salim naek kena kai, rak ngadon sare. Teu wanieun sare di handap, sieun disampeurkeun sato galak, didodoho eukeur sare.& 
        Terekel Salim nack kena kai, rak neadon sare. Teu wanieun sare di handap, sieun disampeurkeun sato galak, didodoho eukeur& 
        Terekel Salim naek kena kai, rak ngadon sare. Teu wanieun sare di handap, sieun disampeurkeun sato galak, didodoho eukeur sare. \\
        \hline
        \includegraphics[width=0.18\textwidth]{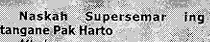} & 
        . eRe ; Naskah Supersemar jing tangane Pak Harto we & 
        Naskah Supersemar ing tangane Pak Harto & 
        Naskah Supersemar ing tangane Pak Harto. \\
        \hline
        \includegraphics[width=0.18\textwidth]{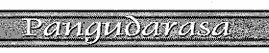} & 
        tn anit ORI ETS .2.Rtee Deira ty cATE SS PEA LIA AROS TA TOS Bie > 2\% NB fe ? 4 oh ry fr any, ct sted ered Sue, ek Acnta, Fil ee na) al Ne la eet tO tee a! aD OP TI PE HOOT Eas & 
        Pangudarasa & 
        Anit ORI ETS. 2. Rtee Deira ty cATE SS PEA LIA AROS TA TOS Bie > 2\% NB fe? 4 oh ry fr any, ct sted ered Sue, ek Acnta, Fil ee na) al Ne la eet tO tee a! aD OP TI PE HOOT Eas. \\
        \hline
    \end{tabular}
    \caption{Comparison of OCR Output, Human Annotation, and Post-OCR Correction with Few-Shot Approach in GPT-4o }
\label{tab:ocr_comparison}
\end{table*}



\begin{table*}[h]
\centering
\begin{tabular}{@{}lcrrr@{}}
\toprule
                       & \multicolumn{4}{c}{\textbf{Languages}}                                                                                                                                \\ \midrule
\multicolumn{1}{l}{} & \multicolumn{1}{c}{\textbf{Balinese}} & \multicolumn{1}{c}{\textbf{Javanese}} & \multicolumn{1}{c}{\textbf{Sundanese}} & \multicolumn{1}{c}{\textbf{Minangkabau}} \\ \midrule
\textbf{GT}                     & \multicolumn{1}{r}{16138}                & 12300& 18558& 30368\\
\midrule
\textbf{OTS}                    & \multicolumn{1}{r}{16207}                & 14471& 18771& 30614\\ 
\textbf{Llama3 (FS)}            & \multicolumn{1}{r}{15955}                & 12897& 18524&30490\\
\textbf{Llama3 (ZS)}            & \multicolumn{1}{r}{16034}                & 11979& 18513&30389\\
\textbf{GPT-4 (FS)}             & \multicolumn{1}{r}{16010}                & 13775& 18641&30232\\
\textbf{GPT-4 (FS)}             & \multicolumn{1}{r}{15941}                & 13123& 18606&29785\\
\bottomrule
\end{tabular}
\caption{Number of tokens retrieved from Indonesian local language archives extraction, comparing between the number of ground-truth (GT), off-the-shelf (OTS) TesseractOCR, Llama3 and GPT-4 both with Few-shot (FS) and Zero-shot (ZS), respectively.}
\label{tab:maxlen-eval}

\end{table*}

\begin{table*}[]
\centering
\begin{tabular}{@{}rlrrrrr@{}}
\toprule
\multicolumn{1}{l}{} &                   & \multicolumn{5}{c}{\textbf{WAR}}                                                                                                                                                                           \\ \midrule
\multicolumn{1}{l}{} & \textbf{Language} & \multicolumn{1}{c}{\textbf{OTS}} & \multicolumn{1}{c}{\textbf{Llama3 (ZS)}} & \multicolumn{1}{c}{\textbf{Llama3 (FS)}} & \multicolumn{1}{c}{\textbf{GPT-4 (ZS)}} & \multicolumn{1}{c}{\textbf{GPT-4 (FS)}} \\
\midrule
1                    & Balinese          & 0.777                            & 0.808                                    & 0.818                                    & 0.795                                   & 0.757                                   \\
2                    & Javanese          & -4.04                            & 0.532                                    & -0.966                                   & -2.080                                  & -3.012                                  \\
3                    & Sundanese         & 0.777                            & 0.903                                    & 0.872                                    & 0.760                                   & 0.787                                   \\
4                    & Minangkabau       & 0.866                            & 0.806                                    & 0.779                                    & 0.879                                   & 0.879                                   \\
\midrule
\multicolumn{2}{c}{\textbf{avg (\%)}}    & -0.405                           & \textbf{0.762}& 0.375                                    & 0.088                                   & -0.147                                  \\ \bottomrule
\end{tabular}
\caption{Percentage of Word Accuracy Rates (WAR) on different extraction techniques. The off-the-shelf (OTS) column means there are no additional steps to repair the extracted text, while the others perform additional post-correction steps involving LLMs. Overall, Llama3 with a Zero-shot approach performs well compared to the other LLM in all languages.}
\label{tab:war-eval}
\vspace{110mm}
\end{table*}

\bibliographystyle{acl_natbib}
\end{document}